\documentclass[mnsc,nonblindrev]{style}

\OneAndAHalfSpacedXI

\usepackage{natbib}
\usepackage[]{algorithm2e}
\usepackage[noend]{algpseudocode}
\usepackage{longtable, multirow}
\makeatletter
\def\BState{\State\hskip-\ALG@thistlm}
\makeatother

 \bibpunct[, ]{(}{)}{,}{a}{}{,}%
 \def\bibfont{\small}%
 \def\bibsep{\smallskipamount}%
 \def\bibhang{24pt}%
\TheoremsNumberedThrough     
\ECRepeatTheorems

\EquationsNumberedThrough    

\begin{document}


\RUNAUTHOR{Zhong et al.}

\RUNTITLE{Decision Making for Seed Variety Selection}

\TITLE{Hierarchical Modeling of Seed Variety Yields and Decision Making for Future Planting Plans}

\ARTICLEAUTHORS{%

\AUTHOR{Huaiyang Zhong\textsuperscript{1}, Xiaocheng Li\textsuperscript{1}, David Lobell\textsuperscript{2}, Stefano Ermon\textsuperscript{3}, Margaret L. Brandeau\textsuperscript{1}}
\AFF{\textsuperscript{1}Department of Management Science and Engineering, Stanford University, Stanford, CA\newline \textsuperscript{2}Department of Earth System Science, Stanford University, Stanford, CA \newline
\textsuperscript{3}Computer Science Department, Stanford University, Stanford, CA 
 \EMAIL{}} 
} 

\ABSTRACT{%
Eradicating hunger and malnutrition is a key development goal of the 21st century. We address the problem of optimally identifying seed varieties to reliably increase crop yield within a risk-sensitive decision making framework. Specifically, we introduce a novel hierarchical machine learning mechanism for predicting crop yield (the yield of different seed varieties of the same crop). We integrate this prediction mechanism with a weather forecasting model, and propose three different approaches for decision making under uncertainty to select seed varieties for planting so as to balance yield maximization and risk. We apply our model to the problem of soybean variety selection given in the 2016 Syngenta Crop Challenge. Our prediction model achieves a median absolute error of 3.74 bushels per acre and thus provides good estimates for input into the decision models. 
Our decision models identify the selection of soybean varieties that appropriately balance yield and risk as a function of the farmer's risk aversion level. More generally, our models support farmers in decision making about which seed varieties to plant.
}%


\maketitle

%


\section{Introduction}
Nearly 800 million people -- one-ninth of the world's population -- go to bed hungry every night and one person in three suffers from some form of malnutrition \citep{wfp}. In the coming decades, this problem is likely to be exacerbated by growing populations, changing climate, and environmental stressors \citep{foodsecurity}. Eradicating hunger and malnutrition is thus one of 17 Global Goals for Sustainable Development adopted by the United Nations in 2015 \citep{UNgoals}. 

\begin{figure}[t]
\begin{center}
\includegraphics[height=3in]{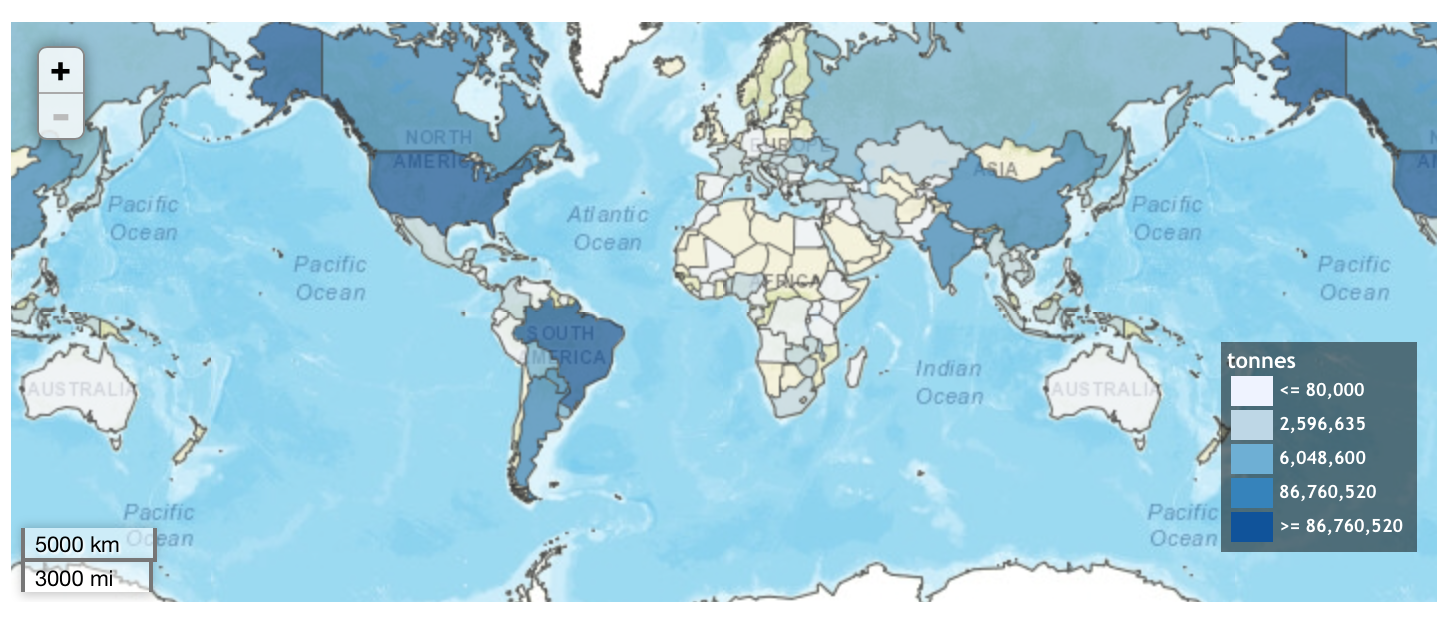}
\caption{2014 Global soybean production} \label{GP}
\end{center}
\end{figure}

Means for improving food security include expanding arable land, increasing cropping intensity, and improving crop yield. This paper focuses on the latter. Specifically, we address the problem of optimally identifying seed varieties to increase crop yield within a risk-sensitive decision making framework. We performed this research as part of the Syngenta Crop Challenge, which focuses on developing and applying innovations in analytics to address the problem of worsening worldwide hunger \citep{ideaconnection}. The goal of the 2016 Syngenta Crop Challenge was to use data on soil properties, seed varieties, and weather patterns to develop a model to determine the soybean varieties that farmers should plant in the next year to reliably reduce risk and increase yields \citep{syngentanews}. Approximately 325 million metric tons of soybeans are grown worldwide each year, with significant amounts of production in North America, South America, and Asia (Figure \ref{GP}). 

The goals of our analysis are twofold: 1) to provide a general method for seed variety selection for seeds of a single crop (e.g., corn, soybeans) -- and thus the development of planting plans -- that takes into account soil and regional information as well as uncertain factors such as weather; 2) to develop a model to support farmers in local decision making about which soybean varieties to plant in the next year (the Syngenta Crop Challenge).

To understand the underlying mechanism of seed variety yield, we construct a two-layer hierarchical model that separates the task of variety yield prediction into two parts: the prediction of check yield (which we define as the average yield of the crop in a site) and the prediction of variety ratio (which we define as variety yield divided by check yield). 

We use machine learning techniques to build and adapt our models. We predict the yield of each seed variety based on available data. To minimize the potential influence of data imbalance (where some seed varieties have many samples of yield but others have few samples), we develop a data augmentation method. The machine learning method combined with the data augmentation method allows us to identify seed varieties that will maximize expected yield, based on knowledge of soil and region attributes. 

We then use an empirical Monte Carlo sampling method to predict yield under weather uncertainty. We also incorporate uncertainty due to other contributing factors, such as seed quality, farming skills, pests, and diseases, which we capture using the residual errors of our variety prediction model. 


Finally, we develop three decision making models for selecting varieties. The models aim to optimize a combination of expected yield and risk. Risk is defined as variation of actual crop yield from expected yield (which is influenced by weather and other sources of uncertainty such as seed quality, farming skills, pests, and diseases). We consider: 1) a utility function model with a risk aversion parameter $\lambda$ chosen by the decision maker; 2) a model that maximizes yield subject to a controlled level of risk $\beta$ determined by the decision maker; and 3) a robust optimization model in which the goal is to maximize the $\alpha$-quantile of yield.

\section{Methodology} \label{Methodology} 


Our model consists of two parts: 1) a \textbf{variety yield prediction scheme}, built using machine learning techniques and 2) a \textbf{decision making under uncertainty framework} for selecting an optimal mix of varieties based on solving an optimization problem that relies on the yield prediction model. 

\subsection{Variety Yield Prediction Model}

\begin{figure}[t]
\begin{center}
\includegraphics[height=3in]{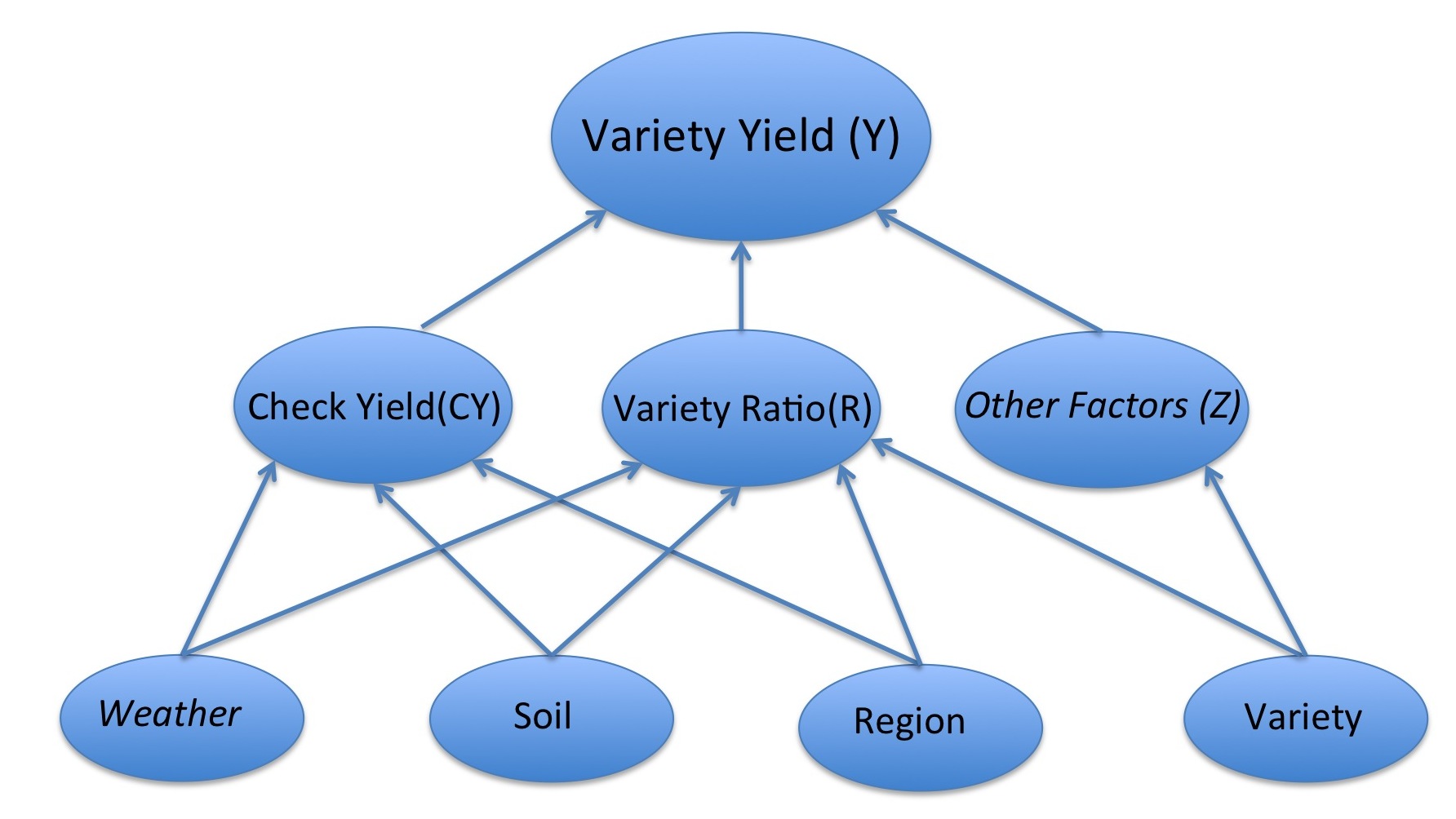}
\caption{Hierarchical yield prediction model. Weather, soil and region attributes are used to predict check yield for a given site (field), as well as variety ratios, which are then combined with other random attributes to produce a final variety yield estimate.} \label{model1}
\end{center}
\end{figure}

We first build a two-layer hierarchical model for variety yield prediction as shown in Figure \ref{model1}. The bottom layer contains four attributes: weather, soil, region and variety.  Notably, soil, region, and variety are fixed attributes, while weather is a random attribute. In the middle layer, to reflect the production level of a given site, we introduce the check yield $CY$ as the average yield of all varieties of the crop in that site. The variety ratio $R$, which is variety yield $Y$ divided by check yield $CY$, reflects the relative expected yield of each variety in that site. However, variety yield $Y$ is also influenced by other factors $Z$ that include effects from all the unknown factors that might contribute to the variety yield but are not measured in the dataset, such as farmer's expertise, quality of seeds, and pests and diseases.

In this hierarchical model, the variety yield $Y$ can be represented as: 
$$Y=CY\cdot R+Z,$$
where
$$CY = F(\text{Weather}, \text{Soil}, \text{Region}),$$
$$R = G_v(\text{Weather}, \text{Soil}, \text{Region}).$$
Here $v$ is the index for seed variety.

During the training phase, we train a single learner $F$ to predict the check yield and variety-specific learners $G_v$ for prediction of variety ratios. At test time, predictions are done in a bottom-up fashion, obtaining a final prediction for variety yield. Weather attributes are regarded as unknown (and modeled as a random variable) when planning for the following year. Additional effects $Z$ not explicitly included in the dataset are also modeled as random variables.

\subsection{Decision Models for Variety Selection} \label{DAmodel} In the yield prediction model, the random variables $Y_v$ represent the variety yield of the $v$-th variety $(v=1,...,N)$. Here $N$ is the total number of seed varieties. We define a decision variable $p = (p_1, ..., p_N)$ which denotes the fraction of each seed variety $v$ that will be planted. We have the constraint that $\Sigma_{v=1}^N p_v = 1$. To choose a variety mix $p$ that balances expected yield with risk of low yield, we formulate the decision making problem using three different models.  We denote the expectation of the yield $Y$ as $\mu$, the covariance matrix of $Y$ as $\Sigma$, and the sets of constraints on $p$ as $\mathcal{C}.$

\begin{itemize}
\item \textbf{Utility function model:}
\begin{equation}
\max_{p} \ \ U = p^T \mu - \lambda p^T \Sigma p
\label{1model}
\end{equation}
$$\text{subject to} \ \ \ p \in \mathcal{C}.$$
We formulate a utility function with respect to the choice of $p$, based on the expected yield and the variance of yield. The goal is to maximize expected crop yield minus a weighted function of the yield variance. The parameter $\lambda$ ($\lambda \ge 0$) reflects risk aversion and is chosen by the decision maker. When $\lambda = 0$, the decision maker is risk indifferent and expected yield is maximized. As $\lambda$ increases above zero, risk aversion increases.

\item \textbf{Controlled-risk yield maximization model:}
\begin{equation}
\max_{p} \ \ p^T \mu 
\label{2model}
\end{equation}
$$\text{subject to} \ \ \ p^T \sum p \le \beta$$
$$p \in \mathcal{C}.$$

\noindent Under this formulation, we maximize the expected crop yield subject to a maximum allowable risk level $\beta$ ($\beta > 0$).

\item  \textbf{Robust optimization model:}
\begin{equation}
\max_{p,t} \ \ y 
\label{3model}
\end{equation}
$$\text{subject to} \ \ \text{Prob}(p^TY\le y)\le \alpha$$
$$p \in \mathcal{C}.$$
\end{itemize}

\noindent This is a robust optimization model in which we treat the yield as a random variable, $y$, and maximize its $\alpha-$quantile ($\alpha \in [0,1]$). In other words, we look for an allocation $p$ such that the farmer can obtain yield $y$ with probability $1-\alpha.$ As $\alpha$ approaches 1, expected yield is maximized. As $\alpha$ decreases, risk sensitivity increases. When $\alpha$ is near zero, the model maximizes the minimum yield that can be obtained.

It is straightforward to convert the first two decision models into integer programs and solve them using a solver such as CPLEX \citep{CPLEX}. The third model is more complicated to solve. For this model we develop an efficient heuristic algorithm, given in the Appendix. The algorithm builds lists of variety combinations in a greedy fashion: First the variety is chosen that maximizes $y$ subject to the quantile constraint. Then an additional variety is added to maximize the incremental improvement in $y$ subject to the constraint. The process continues until no more varieties can be added.

\section{Application to Syngenta Crop Challenge} \label{Quantitativeresults}

In this section, we apply our models to the Syngenta dataset to identify the soybean varieties that should be planted in the next year. The Syngenta dataset that we used contains information on $182$ soybean varieties in $117$ sites over $7$ years (from 2008 to 2014). The dataset provides information on $30$ attributes, listed in Appendix Table B1. The dataset contains approximately $34,000$ entries. 

\subsection{Fixed and Random Attributes for the  Yield Prediction Model} \label{Attr}

In Figure \ref{model1}, the bottom layer includes both fixed attributes (soil, region, variety) and random attributes (weather). Information on each of the fixed attributes is provided in the Syngenta dataset:
\begin{itemize}
\item Soil (16 attributes): sand, silt, clay, pH value, etc.
\item Region (6 attributes): longitude, latitude, probability of growing soybeans, etc.
\item Variety (1 attribute): the variety type index.
\end{itemize}

There are two classes of random attributes: weather and other factors ($Z$). Weather attributes are very important in the yield prediction process. We use an empirical Monte Carlo resampling procedure which utilizes historical weather data and can generate random weather attributes samples based on location, climate, and weather types. All of the other unknown factors that might affect the yield are incorporated into a single attribute ($Z$). This attribute could be interpreted as the noise of variety yield or could be interpreted as explaining the residual errors in our prediction models. We define this attribute as variety-specific based on the observations that yield can vary even in the same site with the same variety and some varieties inherently have a larger yield variance than others.

Thus, the random attributes are:

\begin{itemize}
\item Weather (6 attributes): precipitation, radiation and temperature in the year of experiments (random) and their historical medians (fixed).
\item Other factors (1 attribute): variety-specific random variable.
\end{itemize}
   
We now discuss soybean variety categorization and yield prediction. We use $28$ of the above $30$ attributes -- those relating to soil, region and weather (see Figure 2) -- to categorize the soybean varieties and to predict check yield (for these tasks, we set variety as fixed, and we do not include the variety-specific random variable $Z$ that captures the non-weather random factors).

\subsection{Data Selection, Soybean Variety Categorization, and Data Augmentation} \label{Clustering}

The Syngenta dataset is heavily imbalanced: some varieties have more than 1000 experimental trials while others have far less. Since the prediction model uses 28 attributes, the variety-specific yield ratio learner would be unreliable if trained with fewer than $30$ samples. Therefore we only consider the $80$ most frequent varieties in the dataset. The least commonly used variety in this subset has 30 samples in the training set (according to our cross-validation split).

A large proportion of the 80 candidate varieties have very few samples compared to others. To alleviate this issue, we introduce a data augmentation scheme in order to make the learner more stable and to reduce the test error. The idea is to utilize ``similar" varieties to assist the training process. 

\begin{figure}[t]
\begin{center}
\includegraphics[height=6.0in]{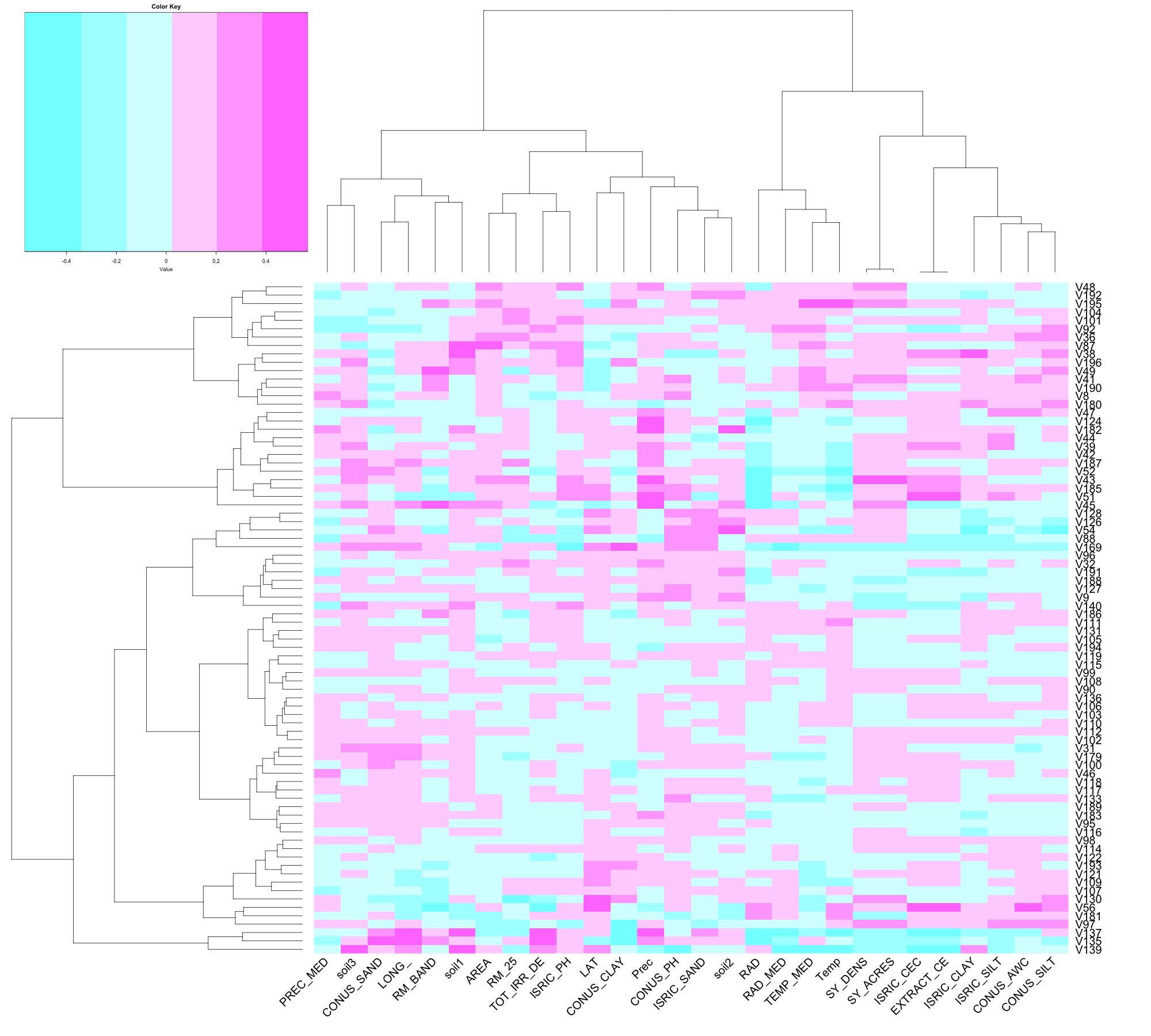}
\caption{Bi-clustering plot for variety yield \scriptsize{\textnormal{Biclustering is conducted on the correlation matrix of variety yield (rows) versus attributes (columns). The intensity of color in the plot reflects the strength of correlation (pink for positive and blue for negative). The closer two rows/columns are, the more similarities the two varieties/attributes share. }} } \label{cluster}
\end{center}
\end{figure}

Figure~\ref{cluster} shows the correlation matrix of attributes and soybean variety yields. If a group of attributes and varieties are clustered horizontally or vertically, there exists a strong similarity among them that can be informative and helpful in practice. Large clusters are present in Figure~\ref{cluster}, indicating strong correlation between attributes and variety yields and suggesting potential categorization of soybean varieties.

A close look at the rows (variety yields) reveals clusters of related attributes. A hierarchical representation of the clusters is provided along the top of Figure \ref{cluster}.  For instance, precipitation (PREC), top soil pH (CONUS\_PH), soil sand content (ISRIC\_SAND) and soil type 2 (soil2) display a similar pattern of correlations with the yield of all varieties. This information is useful because it suggests alternative ways of improving variety yield: for example, one could change attributes such as top soil pH instead of precipitation, a factor that is beyond the farmer's control. 

A hierarchical representation of varieties is shown on the left side of Figure \ref{cluster}. For instance, varieties 191, 188, 127 and 9 represent a potential cluster, as they have similar correlation signs over almost all the attributes. From the color plot, we observe a number of "blocks" in the rightmost and middle columns. This suggests similarities among certain varieties, and can be used to validate our method for grouping soybean varieties.

To obtain a grouping of the varieties, we utilize k-means clustering. We determine the number of categories by plotting the within group sum of squares, as shown in Figure 4. We observe that for $k < 12$, the within group sum of squares continues to decrease as $k$ increases; for $k > 12$, the decrease is relatively small. Based on this we select 12 clusters. The set of soybean varieties in each cluster is shown in Table \ref{AT1}. 

\begin{figure}[t]
\begin{center}
\includegraphics[height=2.4in]{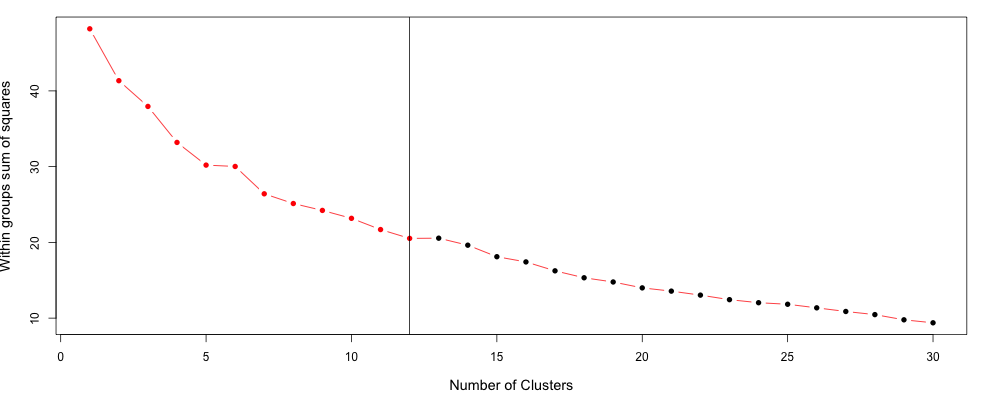}
\caption{Determination of cluster size}
\end{center}
\label{AF1}
\end{figure}

\begin{table}[!h]
\begin{center}
     \begin{tabular}{c|c}
    \hline
  \ Group & Variety \\ \hline
1 &  V56, V181    \\ \hline
2 &  V9, V32, V96, V107, V119, V127, V140, V188, V191    \\ \hline
3 &  V31,  V46,  V90,  V95,  V98,  V99, V100, V102, V103, \\&  V105, V106, V108, V110, V112, V114, V115, V117,\\ & V118, V131, V133, V136, V179, V183, V189, V194  \\ \hline
4 & V88, V116, V126, V128, V192     \\ \hline
5 &  V54    \\ \hline
6 &  V8, V36,  V38,  V41,  V44,  V49, V97, \\& V111, V180, V186, V190, V196   \\ \hline
7 &  V169    \\ \hline
8 &  V139    \\ \hline
9 &  V135, V137   \\ \hline
10 &  V47,  V51, V109, V121, V122, V130, V193     \\ \hline
11 &  V39,  V42,  V43,  V45, V52, V124, V182, V185, V187   \\ \hline
12 &  V48,  V87,  V92, V101, V104, V195  \\  \hline
       \end{tabular}
   \end{center}
   \vspace{.4cm}
   \caption{Cluster Results for K means (K=12) }
   \label{AT1}
 \end{table}

To account for the fact that some of the 80 considered soybean varieties have far fewer samples than others, we use the following data augmentation procedure: during training, if a variety has fewer than 200 samples, we augment the data by randomly sampling from varieties in the same category. Additionally, for the attribute CONUS\_PH (top soil pH), we directly remove those samples for which its value missing. For the attribute RM\_BAND (relative maturity band, a measure of when in the growing season the variety matures), we train a random forest model to predict its value from other attributes and impute the missing values by doing prediction.

We use a 3:1:1 split of the Syngenta data set to create the training, validation and test datasets. All the performance statistics below are reported on the test set.  

\subsection{Yield Prediction}

\subsubsection{Check Yield Prediction}  \label{checkYield}

\begin{figure}[t]
\begin{center}
\includegraphics[height=3in]{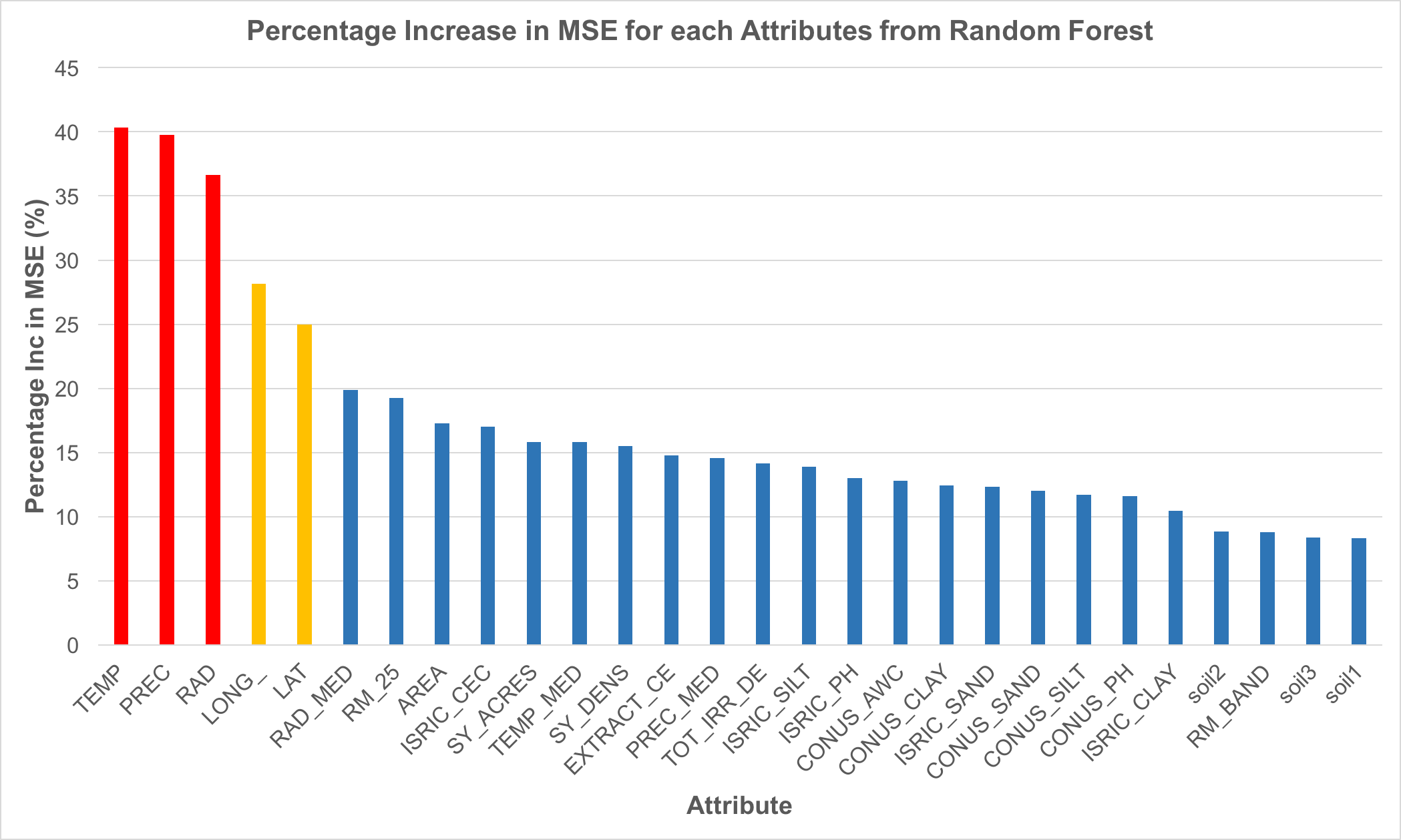}
\vspace{.1cm}
\caption{Attribute importance in check yield prediction \scriptsize{\textnormal{To measure the importance of the $i$-th attribute, we randomly permute the values of $i$-th attribute in the test set and observe the error increase. In this way, we can interpret the attribute importance in check yield prediction.}}}  \label{ImpVar}
\end{center}
\end{figure}

Check yield indicates the current production level of all varieties in a given site, which can be treated as the reference point for a specific site-year combination. In our model (Figure \ref{model1}), predicted check yield serves as a building block for the prediction of variety-specific yield in the same setting. 

For the check yield prediction process we considered several different machine learning models, including linear regression, decision tree models, and neural network models. Based on characteristics of the Syngenta dataset, we choose a random forest model for the check yield prediction. The mean squared errors (MSE) on the test set are reported in Table 2. The baseline model simply uses the mean check yield for all samples as a prediction. With $77.7\%$ of the error explained, the random forest model provides reasonably accurate predictions.

\begin{table}[!h]
   \begin{center}
     \begin{tabular}{c|c|c}
    \hline
 & \ \ \ \ \ Baseline \ \ \ \  \   & Random Forest Model \\ \hline
\ MSE \ &  107.08 & 23.91 \\ 
\hline
       \end{tabular}
   \end{center}
   \label{Check Yield}
   \vspace{.4cm}
   \caption{Mean Squared Errors of Check Yield Prediction}
\end{table} 

To better evaluate performance, we randomly permute the values of each attribute to see how much the check yield prediction error will increase. As shown in Figure \ref{ImpVar}, weather attributes, namely Temperature (Temp), Precipitation (Prec) and Radiation (Rad), increase the mean squared errors (MSE) by $40.35\%$, $39.78\%$ and $36.62\%$, respectively. This result suggests that prediction of Temperature, Precipitation and Radiation for the next growing year is highly significant for predicting the check yield. For this reason we deal with the weather attributes as random variables rather than using a deterministic prediction. Aside from weather attributes, the next two most important attributes are Longitude ($28.13\%$) and Latitude ($24.97\%$) of the sites. The sites with low latitude (towards the equator) and low longitude (towards the West) will tend to have higher check yield. 

\subsubsection{Variety Ratio Prediction}

Variety ratio is the ratio of the expected yield of a given variety in a given site to the average yield of all varieties in the site. Ideally this quantitity is calculated as $\text{variety yield}/\text{check yield}$ (i.e., $R=Y/CY$). The baseline for comparison is to predict all the ratios as $1$, which means simply using the check yield prediction as the variety yield. For each variety, we train a random forest model to estimate its variety ratio. Figure \ref{Ratio} shows the performance of this method on the test set. This is a relatively difficult prediction task: on average, we reduced mean squared error by $20.8\%$ compared to using a variety yield ratio of $1$ for all varieties.

\begin{figure}[h!]
\begin{center}
\includegraphics[height=3.8in]{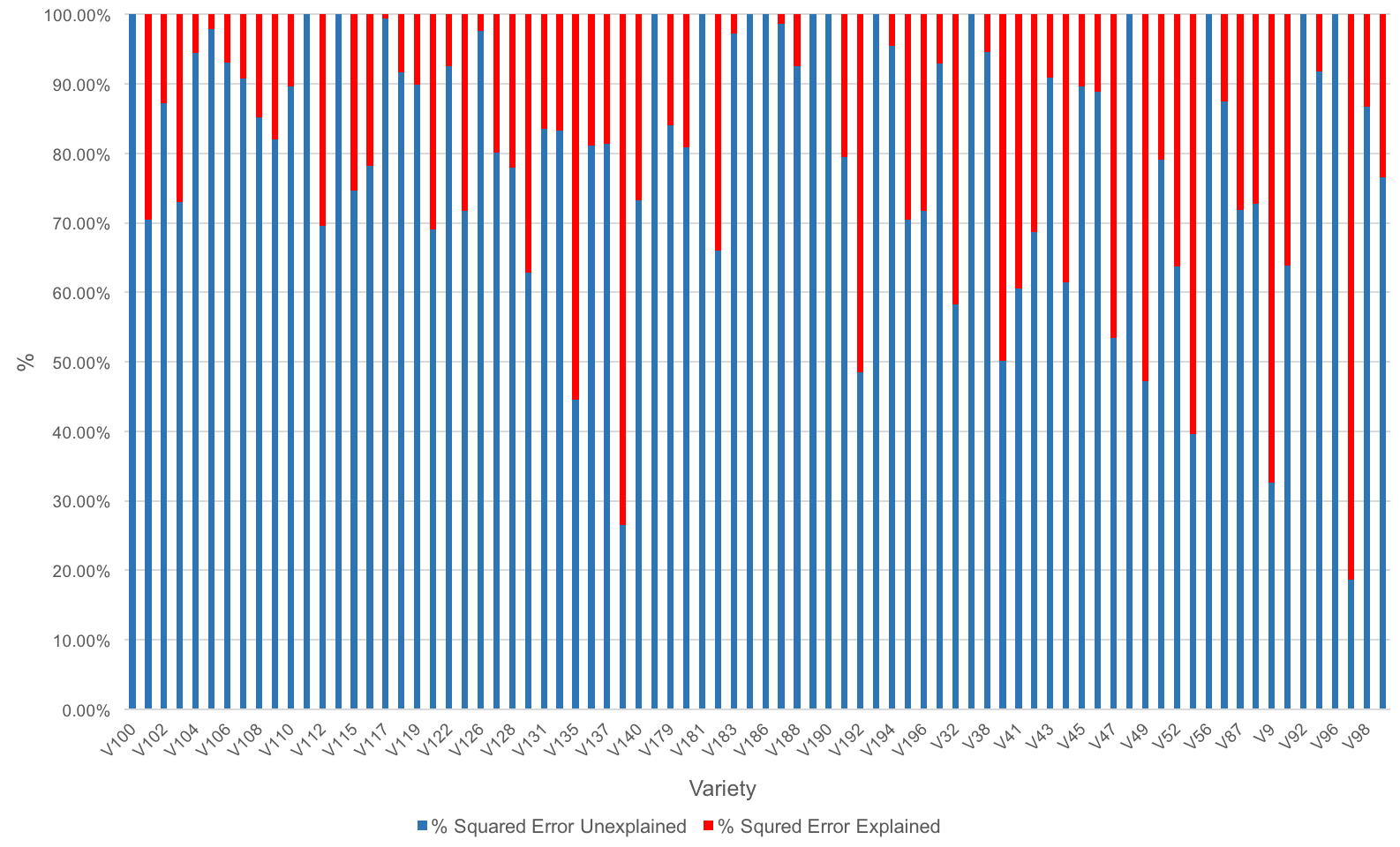}
\caption{Performance of variety ratio prediction  \scriptsize{\textnormal{Each bar of the plot represents a variety; 80 variety ratio predictors are trained. The red bars reflect the error reduction and blue bars represent the residual error.}}} \label{Ratio}
\end{center}
\end{figure}

\subsubsection{Variety Yield Prediction}

By multiplying the prediction of check yield and variety ratio, we obtain a predictor of variety yield (i.e., $Y$ = $CY \times R$). Table \ref{VarietyYield} shows the performance of several variety yield prediction methods. The Baseline model simply predicts the variety yield as the mean yield of all samples. The Check model predicts variety yield as the check yield. The 1-Layer model directly predicts the variety yield with the bottom layer in Figure \ref{model1} (by training variety-specific learners), while the 2-Layer model is the full model shown in Figure \ref{model1}. The 2-Layer DA model additionally employs the data augmentation scheme described above.

\begin{table}[!h]
   \begin{center}
     \begin{tabular}{c|c|c|c|c|c}
    \hline
  &Baseline & Check & One-Layer & Two-Layer & Two-Layer DA \\ \hline
\ MSE \ &  107.08 & 49.74 &45.34 & 41.75 &38.26 \\ \hline
       \end{tabular}
   \end{center}
   \vspace{.4cm}
   \caption{Mean Squared Errors of Variety Yield Prediction Methods \scriptsize{\textnormal{The Baseline model predicts the variety yields as the mean of all yields; the Check model predict variety yield with check yield; the One-Layer model is a single layer model without the middle layer in Figure \ref{model1}; the Two-Layer model reflects the full model in Figure \ref{model1}; and the Two-Layer DA model further utilizes the data augmentation scheme detailed in Section \ref{DA}.}}}
   \label{VarietyYield}
\end{table} 

\begin{figure}[t]
\begin{center}
\includegraphics[height=4in]{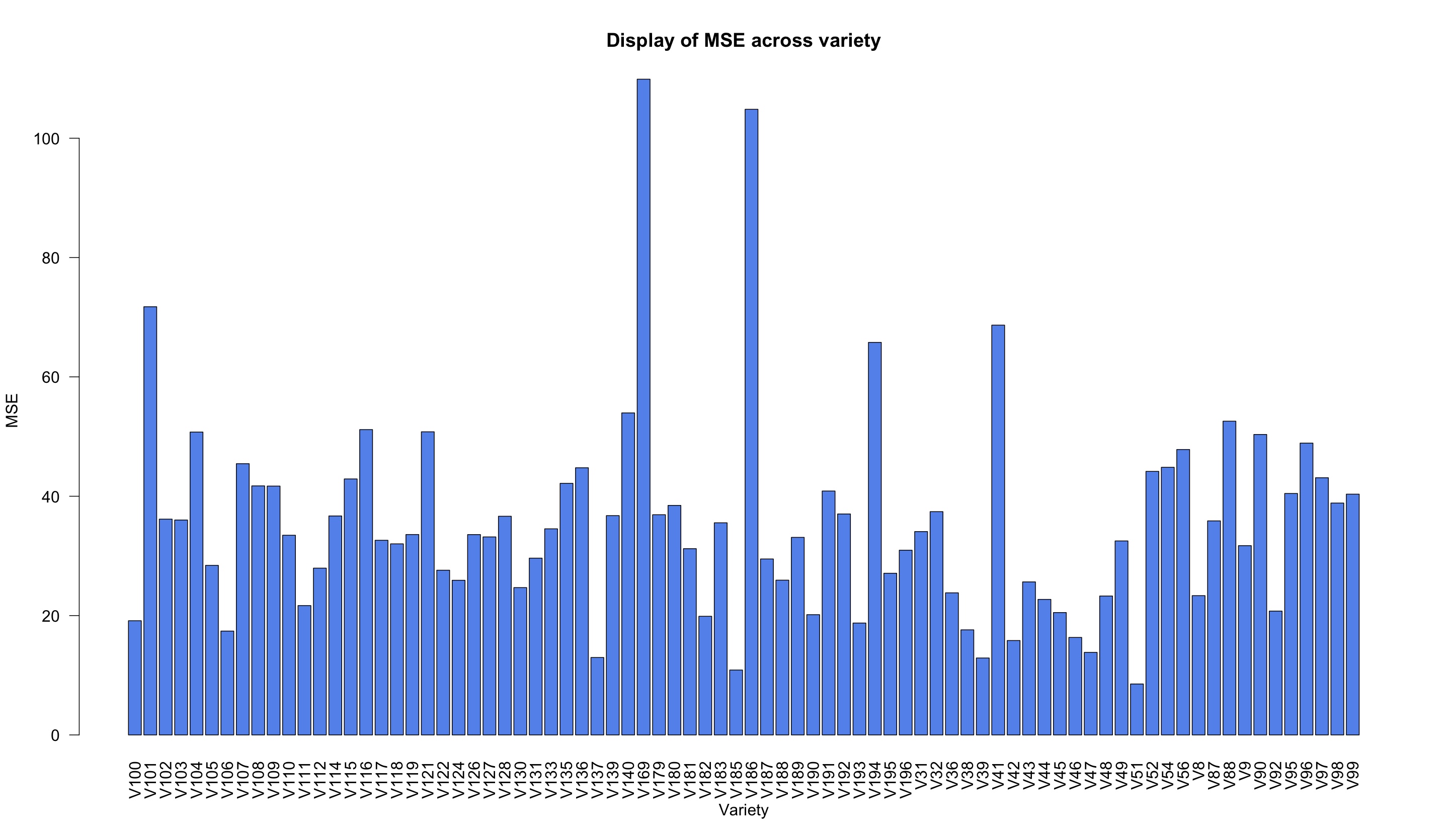}
\caption{Mean squared errors for variety yield prediction \scriptsize{\textnormal{}}} \label{Ratio}
\end{center}
\end{figure}

From Table 3, we can see that the introduction of the middle layer improves the prediction accuracy. Prediction accuracy is further improved by use of our data augmentation scheme, demonstrating its success in dealing with the data imbalance problem. In terms of absolute errors, our model (the Two-layer DA model) obtains a median error of 3.74 bushels per acre and a mean error of 4.70 bushels per acre. In other words, our variety yield prediction on average will lie in a range of $\pm4$ bushels per acre of the true yield. 

As illustrated in Figure 6, the prediction mean squared errors vary significantly by variety types. For some varieties the predictor performed well but for other varieties the residual error is quite large. This phenomenon might be caused by the nature of soybean varieties: some have stable yields while others are more likely to be affected by uncertain factors, including not only weather, but also factors such as farming skills, seed quality, pests and diseases (captured by our random attribute $Z$). In our decision models, we model the random attribute $Z$ by sampling Gaussian noise for the variety yield that is proportional to these MSEs.

\subsection{Variety Selection}


We now use our variety yield prediction model along with information about the uncertain attributes to select soybean varieties to plant in the next year in a given site. This site, in the US midwest, was specified in the Syngenta Crop Challenge. We consider each of the three decision models. The goal is to choose a vector $p$ such that up to $5$ entries of $p$ can be non-zero ($||p||_0\leq5$) with allowable increments of $10\%$ for the fraction of each variety selected.

Since the weather conditions for next year are unknown, we randomly sample weather attributes $w_i$ $(i=1,...,N)$ from historical data of sites around the target site. We created $N=500$ random samples. For each weather attribute $w_i$ and variety type $v_j$, the yield $Y_{ij}$ is calculated by multiplying the check yield prediction and variety ratio prediction. We also add a Gaussian noise proportional to the prediction errors of each variety (as in Figure \ref{Ratio}) to characterize the fact that some varieties are more predictable than others.

Since an autoregressive integrated moving average (ARIMA) model did not yield a parametric fit for the weather time series data (presumably due to weather unpredictability), we design an empirical weather attribute sampling scheme. The idea is to generate samples from historical data on sites that are either near the target site or that share the same climate type as the target site. We first use a Euclidean distance search to identify the 20 nearest neighbors of the target site, and we identify sites that share the same climate type. We accumulate all the historical data from these ``similar" sites and draw random samples from them as weather attribute samples for the next year.

With the empirical random samples $Y$, we can estimate the expected return $\mu$ and $\Sigma$ as the covariance matrix. For the robust optimization model (\ref{3model}), we treat the empirical data $Y$ as the true distribution and maximize the empirical quantiles based on $Y$.

\begin{table}[!h]
   \begin{center}
     \begin{tabular}{c|c|c|c}
    \hline
  $\lambda$ & 0.03 & 0.06 & 0.1 \\ \hline
Selected Varieties &  V41, V44 & V124, V41,V44  & V124, V41, V43, V44   \\ \hline
Combination & $(0.9, 0.1)$ & $(0.2, 0.6, 0.2)$ & $(0.2, 0.5, 0.1, 0.2)$  \\ \hline
Expected Yield (Bushels per Acre) &  59.29 & 58.75 & 58.52 \\ \hline
       \end{tabular}
   \end{center}
   \vspace{.4cm}
   \caption{Decisions for Utility Function Model \scriptsize{\textnormal{The $\lambda$ value reflects the risk aversion level in model (\ref{1model}) in section \ref{DAmodel}.}}}
   \label{M1}
   \vspace{.4cm}
   \begin{center}
     \begin{tabular}{c|c|c|c}
    \hline
  $\beta$ &  100 &  80 & 60 \\ \hline
Selected Varieties &  V41 & V124, V41 &  V124, V41,V44 \\ \hline
Combination & $(1.0)$ & $(0.9, 0.1)$ & $(0.2, 0.6, 0.2)$  \\ \hline
Expected Yield (Bushels per Acre) &  59.40 &  59.19 & 58.75 \\ \hline
       \end{tabular}
   \end{center}
   \vspace{.4cm}
   \caption{Decisions for Controlled-risk Yield Maximization Model
   \scriptsize{\textnormal{The $\beta$ value reflects the risk aversion level in model (\ref{2model}) in section \ref{DAmodel}.}}}
   \label{M2}
     \vspace{.4cm}
 \begin{center}
     \begin{tabular}{c|c|c|c}
    \hline
  $\alpha$ &  0.2 & 0.5  & 0.8 \\ \hline
Selected Varieties &  V41 & V44, V124 & V124  \\ \hline
Combination &  (1.0) & (0.5, 0.5) & 1.0  \\ \hline
$\alpha-$Quantile Yield (Bushels per Acre) &  57.12 & 58.73 & 61.17 \\ \hline
Expected Yield (Bushels per Acre) &  59.40 &  57.78 & 57.31 \\\hline
       \end{tabular}
   \end{center}
   \vspace{.4cm}
   \caption{Decisions for Robust Optimization Model 
    \scriptsize{\textnormal{The $\alpha$ value is the quantile yield that we aim to maximize in model (\ref{3model}) in section \ref{DAmodel}.}}}
     \vspace{.4cm}
     \label{M3}
 \end{table}

Results from each of our three decision models are presented in Tables \ref{M1}, \ref{M2}, and \ref{M3}, respectively. For a low level of risk aversion, all three models favor variety V41 (the utility function model also includes $10\%$ of variety V44). As the risk aversion level increases (moving toward the rightmost column in each table), the optimal choice for all three models becomes more conservative: a combination of varieties becomes optimal, rather than just the variety with the largest expected yield. With a higher level of risk aversion, all three models include variety V124. The utility function and controlled-risk yield maximization models also include varieties V41 and V44, and the utility function model additionally includes V43. As risk aversion increases, expected yield decreases slightly. We note that, from our k-means cluster analysis (Table \ref{AT1}), varieties V41 and V44 were categorized in the same cluster, and varieties V43 and V124 were categorized together in a different cluster.

\begin{figure}[t]
\begin{center}
\includegraphics[height=4in]{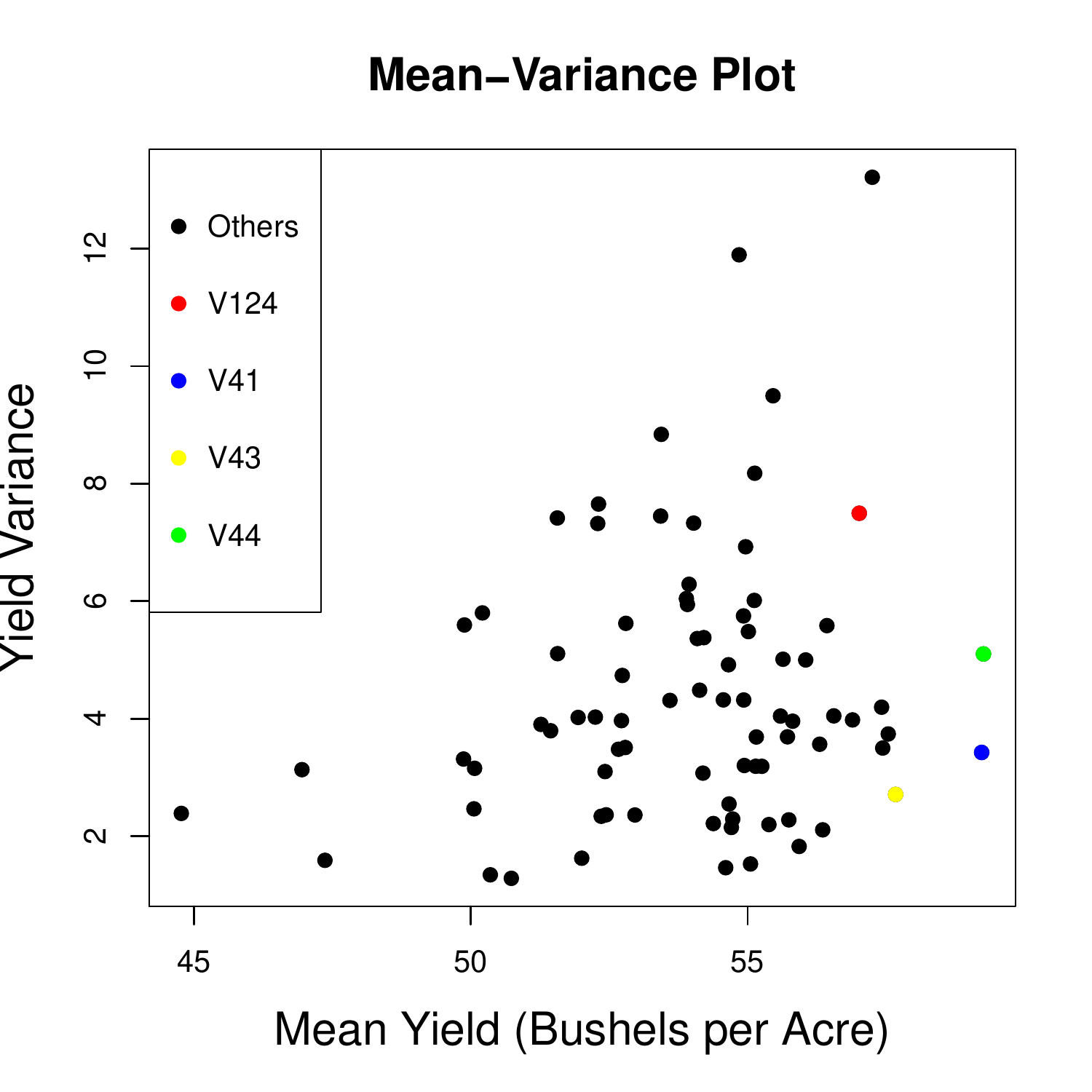}
\caption{Mean-variance plot of all soybean varieties \scriptsize{\textnormal{The x-axis is each variety's expected yield and the y-axis is each variety's yield variance. Each data point represents a variety. The colored points are the varieties selected by the different decision models.}}}

\label{allselection}
\end{center}
\end{figure}

The yield risk is hedged by combining several varieties. The three decision models selected various combinations of the varieties V41, V43, V44 and V124. Figure \ref{allselection} plots the mean and variance of yield for all 80 soybean varieties that we considered. We see that the decision models selected varieties from the right bottom corner where points have high yield and relatively low variance. For a low level of risk aversion, the models select variety V41, which has the highest expected yield of all varieties. As risk aversion increases, the models select variety V44, which has almost the same yield as variety V41 but higher variance; then variety V124, which has lower yield and higher variance; and finally variety V43 (selected by the utility function model) which has lower yield but also lower variance.

After comparing the results from the three decision models, \textbf{we recommend V124, V41 and V44} with a proportion of $(0.2,0.6,0.2)$ as the planting plan for this site for next year.

\section{Discussion} \label{Conclusion}

In this paper, we address the problem of decision making for seed variety selection. Our hierarchical approach combining machine learning with a weather forecasting procedure allows us to accurately predict the yield of different seed varieties in a given site. We develop three decision models for selecting varieties that balance yield maximization and risk given the farmer's risk aversion level. Our prediction and decision models support farmers in decision making about which seed varieties to plant in the next year.

For the Syngenta dataset, our prediction model generated estimates of yield that were within $\pm4$ bushels per acre (or approximately $7\%$) of the true yield. The three decision models identified similar varieties for similar levels of risk aversion, thus providing a clear direction for variety selection.

We have focused on the problem of seed variety selection. However, our modeling approach provides additional information that can support farmers with other decisions. For example, our prediction method identifies how soil attributes affect crop yield. This can be used as a basis for decisions regarding soil amendment. Our prediction method also identifies how weather, soil and region attributes affect yield. This information can be used to assist in selecting promising sites for future farmland. Our clustering method provides information about seed varieties that have similar yield characteristics in a given site. This information can be used heuristically to select alternate seed varieties for a given site (for example, if a particular variety is not available).

Data analytics, simulation, and optimization have been successfully applied in a variety of business applications. Here we apply these techniques to a critical problem in agriculture. Our work provides a simple and effective way to reliably increase crop yield, which in turn can help to reduce hunger and malnutrition around the world. 



\newpage
\bibliographystyle{ormsv080}
\bibliography{sample.bib}

\newpage
\begin{APPENDICES}
\section{Algorithm for Solving Robust Optimization Model}
We use the following heuristic algorithm to efficiently solve the robust optimization model:
\begin{algorithm}[H]
 \KwData{500 runs of variety yield prediction for each variety}
 \KwResult{Selected Variety List,  Maximum $\alpha$\ Quantile Yield } 
 initialization\;
 \While{selected varieties no larger than 5}{
  Empty Selected Variety List \;
  Set Maximum $\alpha$\ Quantile Yield to 0\;
  \eIf{have not selected any variety}{
   find the maximum of $\alpha$\ quantile of 80 variety yields over 500 runs \;
   update Maximum $\alpha$\ Quantile Yield by value obtained\;
   add the variety to the Selected Variety List\;
    
   }{
   find maximum of $\alpha$\ quantile for combinations of selected variety/varieties and each remaining variety\;
   \eIf{Maximum $\alpha$\ Quantile Yield less than above obtained value }
   {update Maximum $\alpha$\ Quantile Yield by value obtained\;
   add the variety to the Selected Variety List\;}
   {Jump out of the while loop}
  }
  }
  
  exhaustive search for all combinations in Selected Variety\;
  find the combination with maximum $\alpha$\ quantile yield\;
\end{algorithm}

\section{Feature Code Key}
Table B1 provides a detailed description of the $30$ features in the Syngenta dataset. 

\begin{longtable}{c|c}
\hline

Feature(s)         & Description\\
\hline
YEAR           & Year\\
SITE             & Site code\\
BREEDING\_G      & Breeding group\\
EXP              & Experiment number\\
LAT              & Latitude\\
LONG             & Longitude\\
VARIETY          & Variety code\\
VARIETY\_YIELD   & Variety yield\\
CHECK\_YIELD     & Check yield for the trial\\
YIELD DIFFERENCE & Yield difference between variety and check in a trial \\
SITE YIELD      & Average site yield (check yields across experiments)\\
RM\_BAND         & Relative maturity band\\
CLIMATE         & Climate type (based on temperature, precipitation and solar radiation)            \\
SEASON           & Season type \\
AREA             & Probability of growing soybeans\\
RM\_25           & Probability of growing soybeans of relative maturity band 2.5 to 3 \\
TOT\_IRR\_DE     & Probability of irrigation \\
SOIL\_CUBE       & Soil type (based on texture, available water holding capacity, soil drainage) \\
TEMP\_08, ..., TEMP\_14         & Sum of the temperatures for each season, 2008, ..., 2014 \\
TEMP\_MED        & Median sum of temperatures for seasons from 2001 to 2014                     \\
PREC\_08, ... PREC\_14         & Sum of the precipitation for each season, 2008, ..., 2014\\
PREC\_MED        & Median sum of precipitation for seasons from 2001 to 2014                    \\
RAD\_08, ..., RAD\_14          & Sum of the solar radiation for each season, 2008, ..., 2014 \\
RAD\_MED         & Median sum of solar radiation for seasons from 2001 to 2014                  \\
CONUS\_PH        & Topsoil pH (10 to 20 cm depth)  \\
CONUS\_AWC       & Topsoil available water capacity in 150 cm soil profile (10 to 20 cm depth) \\
CONUS\_CLAY      & Topsoil clay content (10 to 20 cm depth) \\
CONUS\_SILT      & Topsoil silt content (10 to 20 cm depth) \\
CONUS\_SAND      & Topsoil sand content (10 to 20 cm depth) \\
ISRIC\_SAND      & Soil sand content from another soil source \\
ISRIC\_SILT      & Soil silt content from another soil source \\
ISRIC\_CLAY      & Soil clay content from another soil source\\
ISRIC\_PH        & Soil pH from another soil source \\
ISRIC\_CEC       & Soil cation exchange from another soil source \\
\hline
\caption*{Table B1: Description of Features}
\label{FeatureKey}
\end{longtable}
\end{APPENDICES}
\end{document}